%% file: main.tex
\documentclass[letterpaper, 10pt, conference]{ieeeconf}
\IEEEoverridecommandlockouts
\input{preamble}

\title{\fontsize{17pt}{24pt}\selectfont \bf Fast Asymptotically Optimal Kinodynamic Planning via Vectorization}

\newif\ifanonymous
\anonymousfalse

\ifanonymous
  \author{Anonymous Author(s)}
\else
    \author{
    Yitian Gao, Andrew Lu, and Zachary Kingston%
    \thanks{YG, AL, and ZK are with the Department of Computer Science, Purdue University, {\tt \{gao634, lu987, zkingston\}@purdue.edu}. 
    }}
\fi

\begin{document}
\maketitle
\thispagestyle{empty}
\pagestyle{empty}

\begin{abstract}
Sampling-based motion planners have been shown to be effective for systems with complex kinodynamic constraints and high dimensionality.
However, these algorithms struggle to achieve real-time performance, leading to recent efforts to parallelize planning.
While GPU-accelerated planners have achieved significant speedups, existing approaches require specialized CUDA programming that limits accessibility and portability. 
We present \textit{Parallel Asymptotically Optimal Kinodynamic RRT} (PAKR), a massively parallel kinodynamic planner leveraging JAX and the XLA compiler to achieve GPU acceleration through standard Python tooling.
By combining our parallel planner with the AO-\textit{x} meta-algorithm, we achieve asymptotic optimality through fast iterative replanning.
We provide a theoretical analysis of probabilistic completeness, analyze the effects of batch size and branching factor on convergence, and demonstrate scalability to complex dynamics using the MuJoCo-XLA simulator.
Experiments show competitive runtimes with state-of-the-art GPU planners and superior solution quality.
\end{abstract}

\section{Introduction}
\label{sec:intro}

Robotic systems deployed in dynamic environments require fast, reactive motion planning that accounts for the robot's dynamics. 
One approach is kinodynamic motion planning, which is concerned with finding trajectories that satisfy both kinematic constraints and equations governing the robot's dynamics, and can be quite challenging for even low-dimensional problems~\cite{lavalle2006planning, kavraki2016motion}.
Sampling-based motion planners (SBMPs) have proven effective due to their ability to handle high-dimensional state spaces and complex nonlinear dynamics~\cite{rrt, sst,orthey2023sampling}.
Still, finding solutions can take hundreds of milliseconds for simple systems and tens of seconds for complex nonlinear dynamics, which is insufficient for real-time reactivity in changing environments.

Recent advances in parallel computing offer a solution.
GPU and CPU parallelization have demonstrated significant speedups by batching operations such as collision checking~\cite{vamp, 6095053}, parallelizing entire expansion iterations~\cite{prrtc, kinopax}, or running multiple independent planner instances~\cite{caselli99, cforest}. 
For kinodynamic systems specifically, Kino-PAX~\cite{kinopax} achieves real-time performance through GPU-accelerated tree expansion, finding solutions in tens of milliseconds.

However, parallel motion planning algorithms often compromise on solution quality.
Many asymptotically optimal (AO) planners achieve high-quality solutions through mechanisms like rewiring~\cite{rrtstar, kinorrtstar} or maintaining best-cost nodes within witness regions~\cite{sst}.
These techniques rely on ordering; each node's insertion depends on the current global state of the tree.
In a batched parallel context, samples within the same batch lack mutual visibility: a node may attach to a sub-optimal parent because the optimal parent is still being processed.
This staleness problem leads to redundant exploration and sub-optimal decisions, making it difficult to maintain theoretical convergence guarantees without introducing significant synchronization overhead.

We present \methodname, a massively parallel kinodynamic motion planner that achieves both speed and solution quality.
Our key insight is that the AO-\textit{x} meta-algorithm~\cite{aox}, which transforms any probabilistically complete planner into an asymptotically optimal one through iterative cost-bounded replanning, is naturally suited to parallelization; rather than fighting stale information, we embrace fast, sub-optimal planning and use rapid replanning to converge toward optimal solutions.
Our implementation uses JAX~\cite{jax2018github} to fuse the entire planning loop into a single GPU kernel, eliminating CPU-GPU overhead without requiring CUDA programming.
Our approach makes GPU-accelerated kinodynamic planning accessible through standard Python tooling.

In summary, our contributions are: a massively parallel kinodynamic SBMP that leverages XLA compilation via JAX for efficient GPU execution, an implementation of the AO-\textit{x} algorithm to achieve asymptotic optimality in a parallel setting, demonstration of scalability to complex domains including MuJoCo-XLA (MJX)~\cite{mujoco}, and experimental validation showing competitive or superior performance compared to state-of-the-art planners, with solutions found in milliseconds.
Our code is available open source at~\url{https://github.com/CoMMALab/pakr}.

\section{Related Work}
\label{sec:lit}

Geometric sampling-based motion planners (SBMPs)~\cite{lavalle2006planning,orthey2023sampling,prm} build a tree or graph of configuration states connected by collision-free motions. 
A configuration state $x_{rand}$ is first sampled in the state space, and the planner attempts to connect this state to the nearest tree or graph node $x_{near}$ via interpolation. 
Several modifications to the RRT algorithm have been proposed to improve exploration and convergence~\cite{kuffner2000rrtconnect, rrtstar, est}.

\subsection{Kinodynamic Planning}

In kinodynamic planning, interpolating between $x_{rand}$ and $x_{near}$ requires solving a Two-Point Boundary Value Problem (BVP): finding a control trajectory $u(t)$ such that the system state $x(t)$ exactly satisfies $x(0) = x_{near}$ and $x(T) = x_{rand}$ while obeying the dynamics $\dot{x} = f(x, u)$. For high-dimensional or non-linear systems, such as those with underactuated dynamics or non-holonomic constraints, the BVP is often computationally intractable or has no closed-form solution. To avoid the BVP in bidirectional search, Nayak and Otte~\cite{gbrrt} use the reverse tree as a heuristic to guide the forward expansion, but do not directly connect the forward and reverse trees.

Sampling-based kinodynamic planners~\cite{rrt, sst, gbrrt} avoid this by adding new states through forward propagation from $x_{near}$, ensuring dynamic feasibility while still guiding expansion towards unexplored regions by selecting $x_{near}$ to be close to $x_{rand}$. Forward propagation can either be guided by a steering function to ensure the new state is close to $x_{rand}$, or it can be random, sampling a control from a uniform distribution~\cite{rrt}.
By bypassing the need for a ``steering" mechanism, kinodynamic planners excel in planning for systems with complex physics, such as agile drones, soft robots~\cite{vine}, and non-prehensile manipulation tasks where the contact dynamics are non-linear and discontinuous~\cite{10052750}.

\subsection{Asymptotically Optimal Motion Planners}

While some motion planners emphasize fast initial solutions, others prioritize the quality of solutions as they converge toward an optimal path. A motion planner is defined as asymptotically optimal (AO) if, provided an optimal solution exists, the probability that the cost of the returned solution approaches the theoretical optimum equals 1 as the number of samples approaches infinity~\cite{gammell2021asymptotically}.

Many AO and near-optimal variants of the RRT algorithm achieve this property by maintaining a tree structure that is dynamically reorganized as better paths are discovered.
RRT*~\cite{rrtstar} introduced the concept of ``rewiring," where new samples look within a local radius to see if they can provide a lower-cost path to existing nodes. This was also extended to the kinodynamic domain, though the requirement to solve the two-point BVP for every rewiring attempt limits its practical use to systems with linear dynamics~\cite{kinorrtstar}.

As an AO kinodynamic planner for general non-linear systems without a BVP solver, Stable-Sparse RRT*~\cite{sst} maintains witness regions that only consider best-cost nodes within each region for expansion, maintaining a sparse set of nodes that improve in cost with each iteration.

Search-based planners typically construct a graph through a lattice decomposition~\cite{lattice-search} and leverage heuristics to find an optimal path. 
For certain classes of dynamics, solutions found on the lattice are provably within a tolerance $\epsilon$ of the continuous optimum~\cite{donald1993,donald1995provably}.
Meanwhile, Iterative Discontinuity Bounded A* (iDb-A*) grows a set of locally optimal trajectories which asymptotically cover the state space and allow for ``discontinuity-bounded'' connections between states that are sufficiently close.
This allows the planner to treat a sequence of dynamically feasible trajectories as a continuous path, which is further refined through trajectory optimization.

Finally, optimality can be achieved through meta-algorithmic wrappers like AO-\textit{x}~\cite{aox}. This approach allows any feasible (but sub-optimal) kinodynamic planner to be transformed into an AO planner. By iteratively decreasing a cost threshold based on the best solution found so far and rejecting any future samples that exceed this threshold, the algorithm forces the underlying planner to produce increasingly efficient trajectories over time. 

\subsection{Parallel Motion Planning}
Parallel motion planners are generally categorized into coarse-grained and fine-grained methods. 
Coarse-grained parallelization typically involves running multiple independent planner instances;
the simplest approach runs separate planners simultaneously and returns the first solution found~\cite{caselli99}. 
More sophisticated frameworks, such as C-FOREST~\cite{cforest}, use a higher-level manager to periodically share the promising paths found.

Fine-grained parallelization targets specific, computationally expensive sub-processes. 
VAMP~\cite{vamp} uses Single Instruction, Multiple Data (SIMD) batching to parallelize collision checking and forward kinematics solving, achieving sub-millisecond runtimes. 
Similarly, Kino-PAX~\cite{kinopax} and pRRTC~\cite{prrtc} parallelize the expansion iteration itself, considering multiple states for tree insertion simultaneously.

Parallelization often conflicts with the mechanisms required for asymptotic optimality.
Algorithms like RRT*~\cite{rrtstar} and SST*~\cite{sst} rely on a strict sequential ordering; each node's insertion is conditioned on the current ``global" state of the tree to determine optimal connection or pruning.
We also note Kino-PAX$^+$~\cite{perrault2026kino}, a recent extension to Kino-PAX that provides asymptotic near-optimality guarantees by only expanding from the best-cost nodes in witness regions, similar to~\cite{sst}.
In contrast, the AO-\textit{x} meta-algorithm~\cite{aox} builds off of parallelism directly: the faster a feasible solution can be found, the more the solution can be improved.

\section{Problem Formulation}
Consider a robot with state space $\mathcal{X} \subseteq \mathbb{R}^n$ and control space $\mathcal{U} \subseteq \mathbb{R}^m$. The system dynamics are governed by:
\begin{equation}
\dot{x}(t) = f(x(t), u(t))
\label{eq:dynamics}
\end{equation}
where $x(t) \in \mathcal{X}$ and $u(t) \in \mathcal{U}$ are the state and control at time $t$. We assume $f$ is Lipschitz continuous in both arguments.
A trajectory $\sigma: [0, T] \to \mathcal{X}$ is \emph{feasible} if there exists controls $u: [0, T] \to \mathcal{U}$ such that $\dot{\sigma}(t) = f(\sigma(t), u(t))$ and $\sigma(t) \in \mathcal{X}_{free}$ for all $t \in [0, T]$, where $\mathcal{X}_{free}$ is the collision-free subset of $\mathcal{X}$.

Let $\Sigma$ denote the set of all feasible trajectories.
We define a cost function $J: \Sigma \to \mathbb{R}_{\geq 0}$ that assigns a non-negative cost to each trajectory.
We assume $J$ satisfies the standard properties for optimal planning~\cite{aox}: Lipschitz continuity, additivity, and monotonicity. 
Common examples include path length $J(\sigma) = \int_0^T \|\dot{\sigma}(t)\| dt$ and time optimality $J(\sigma) = T$.

We are concerned with the problem of optimal kinodynamic planning.
That is, given dynamics $f$, initial state $x_{init} \in \mathcal{X}_{free}$, goal region $\mathcal{X}_{goal} \subseteq \mathcal{X}_{free}$, and cost function $J$, find a control trajectory $u: [0, T] \to \mathcal{U}$ that minimizes $J(\sigma)$ over all feasible trajectories $\sigma$ satisfying:
\begin{equation*}
    \sigma(0) = x_{init}, \quad \sigma(T) \in \mathcal{X}_{goal}, \quad \sigma(t) \in \mathcal{X}_{free} \;\; \forall t \in [0, T].
\end{equation*}

Let $J^* = \inf_{\sigma \in \Sigma_{sol}} J(\sigma)$ denote the optimal cost, where $\Sigma_{sol} \subseteq \Sigma$ is the set of goal-reaching feasible trajectories. An algorithm is \emph{asymptotically optimal} if the cost of its returned solution converges to $J^*$ almost surely as computation time increases~\cite{gammell2021asymptotically}.
\section{Methodology}
\label{sec:meth}

Our key insight is that the AO-\textit{x} meta-algorithm~\cite{aox} is naturally suited to massively parallel execution.
While AO-\textit{x} transforms any probabilistically complete planner into an asymptotically optimal one through iterative cost-bounded replanning, its practical utility depends on the speed of the inner planner.
Typical planners require seconds per iteration, limiting the number of refinement cycles achievable.
Rather than fighting synchronization issues within a single tree, we embrace fast sub-optimal planning and use rapid replanning to converge toward optimality.
By using our parallel kinodynamic RRT (\cref{sec:parallel_rrt}), each inner planning iteration completes in milliseconds, enabling many cycles of improvement.

\begin{algorithm}
\caption{Parallel AO Kinodynamic RRT (\methodname)}
\label{alg:ao_rrt}

\KwIn{Initial state $x_{init}$, Goal region $\mathcal{X}_{goal}$, Control space $\mathcal{U}$, Batch size $B$, Branching factor $A$}
\KwOut{Best trajectory $\sigma^*$ found within time budget}

$c_{best} \leftarrow \infty$\; \label{line:cinit}
$\sigma^* \leftarrow \emptyset$\;
\While{time budget not exceeded}{ \label{line:outer}
    \tcp{Initialize new tree for this planning iteration}
    \tcp{Tree stores (state, action, cost-to-come) tuples}
    $\mathcal{T} \leftarrow \{(x_{init}, \emptyset, 0)\}$\;
    \While{goal not reached}{ \label{line:inner}
        \tcp{Parallel batch expansion}
        $X_{rand} \leftarrow \text{SampleStates}(B/A)$\; \label{line:sample}
        \tcp{Select $B/A$ parents via nearest neighbor}
        $X_{near} \leftarrow \text{NearestNeighbor}(\mathcal{T}, X_{rand})$\;
        \tcp{Sample $A$ controls per parent}
        $U \leftarrow \text{SampleControls}(B/A, A)$\;
        \tcp{Parallel rollout of all $B$ expansions}
        $X_{new}, C_{new} \leftarrow \text{Propagate}(X_{near}, U, \Delta t)$\; \label{line:prop}
        \ForEach{$(x_{new}, c_{new}) \in (X_{new}, C_{new})$ \textbf{in parallel}}{ \label{line:insert}
            \tcp{Cost-bounded pruning}
            \If{$c_{new} + h(x_{new}) < c_{best}$}{ \label{line:prune}
                \If{$\text{CollisionFree}(x_{near} \rightsquigarrow x_{new})$}{
                    $\mathcal{T} \leftarrow \mathcal{T} \cup \{(x_{new}, u, c_{new})\}$\;
                    \If{$x_{new} \in \mathcal{X}_{goal}$}{ \label{line:goal}
                        $c_{best} \leftarrow c_{new}$\;
                        $\sigma^* \leftarrow \text{ExtractPath}(\mathcal{T}, x_{new})$\;
                        \tcp{Exit inner loop to refine with tighter bound}
                        \textbf{break}\;
                    }
                }
            }
        }
    }
}
\Return $\sigma^*$\;
\end{algorithm}

\cref{alg:ao_rrt} presents the parallel AO kinodynamic-RRT (\methodname) algorithm.
The outer loop (Line~\ref{line:outer}) iteratively refines solutions by restarting the planner with progressively tighter cost bounds.
The inner loop (Line~\ref{line:inner}) implements parallel kinodynamic RRT expansion: each iteration samples $B/A$ target states (Line~\ref{line:sample}), finds their nearest neighbors in the tree, samples $A$ random controls per parent, and propagates all $B$ state-action pairs in parallel (Line~\ref{line:prop}).
Nodes are only inserted if their f-cost $f(n) = g(n) + h(n)$ is below the current best solution cost, where $g(n)$ is the cost-to-come and $h(n) = \|x_n - x_{goal}\|_2$ is the Euclidean heuristic (Line~\ref{line:prune}).
This focuses computation on regions that can potentially improve the solution.
When a goal is reached (Line~\ref{line:goal}), the algorithm records the solution, updates $c_{best}$, and restarts with the tighter bound.

\subsection{Implementation}
\label{sec:parallel_rrt}

To maximize performance, our implementation employs a static-graph architecture optimized for GPU execution.
Unlike CUDA-based planners that rely on dynamic branching and host-side management, a statically compiled planner runs entirely within a single GPU kernel, eliminating CPU-GPU device transfer overhead.

To enable the compiler to generate a high-efficiency monolithic kernel, the algorithm imposes a static shape constraint: the maximum tree size ($N$) and the parallel batch size ($B$) must be defined at compile-time for fixed memory offsets.
The search tree is represented as a structure of arrays, where four contiguous buffers store the tree data: $\textbf{states} \in \mathbb{R}^{N \times d}$ for state vectors, $\textbf{actions} \in \mathbb{R}^{N \times m}$ for control inputs, $\textbf{parents} \in \mathbb{Z}^N$ for parent indices enabling path reconstruction, and $\textbf{costs} \in \mathbb{R}^N$ for cost-to-come values.
Node insertion uses indexed slice updates that write batches of new nodes at the current tree frontier.
A scalar counter tracks the number of valid nodes, enabling masked operations over padded arrays.

The computational cost of nearest neighbor (NN) search is often the bottleneck for planners that construct large trees~\cite{nnbottleneck}.
Fortunately, the distance calculations of the NN search are easily vectorizable, allowing it to scale efficiently with GPU core counts. 
To further reduce the NN bottleneck, we introduce a Branching Factor ($A$). Rather than performing a costly NN search for every candidate expansion, we select $K = B/A$ parent nodes and sample $A$ random controls per parent. 
Our experiments show that moderate branching factors ($A \in [32, 128]$) effectively balance exploration efficiency with NN overhead.
To minimize the performance penalty of padding in a growing tree, we employ a tiered nearest neighbor strategy.
We define memory tiers at tree sizes in the range $4^{[6, 10]}$, with specialized kernels for each tier that process only the active portion of the tree. 

Our framework supports arbitrary dynamics through modular propagator functions. 
For example, we can interface directly with MuJoCo-XLA~\cite{mujoco}, enabling planning directly on the differentiable simulator's state representation.
Forward propagation integrates the system dynamics using fourth-order Runge-Kutta integration.
The entire rollout is vectorized, which compiles efficiently.
Each propagation applies the same control for $T$ timesteps, with validity checked at each step via batched collision queries.

\begin{figure*}[t]

     \centering
     \begin{subfigure}[b]{0.32\textwidth}
         \centering
         \adjincludegraphics[width=\textwidth, trim={0.1\width} 0 {0.1\width} 0, clip]{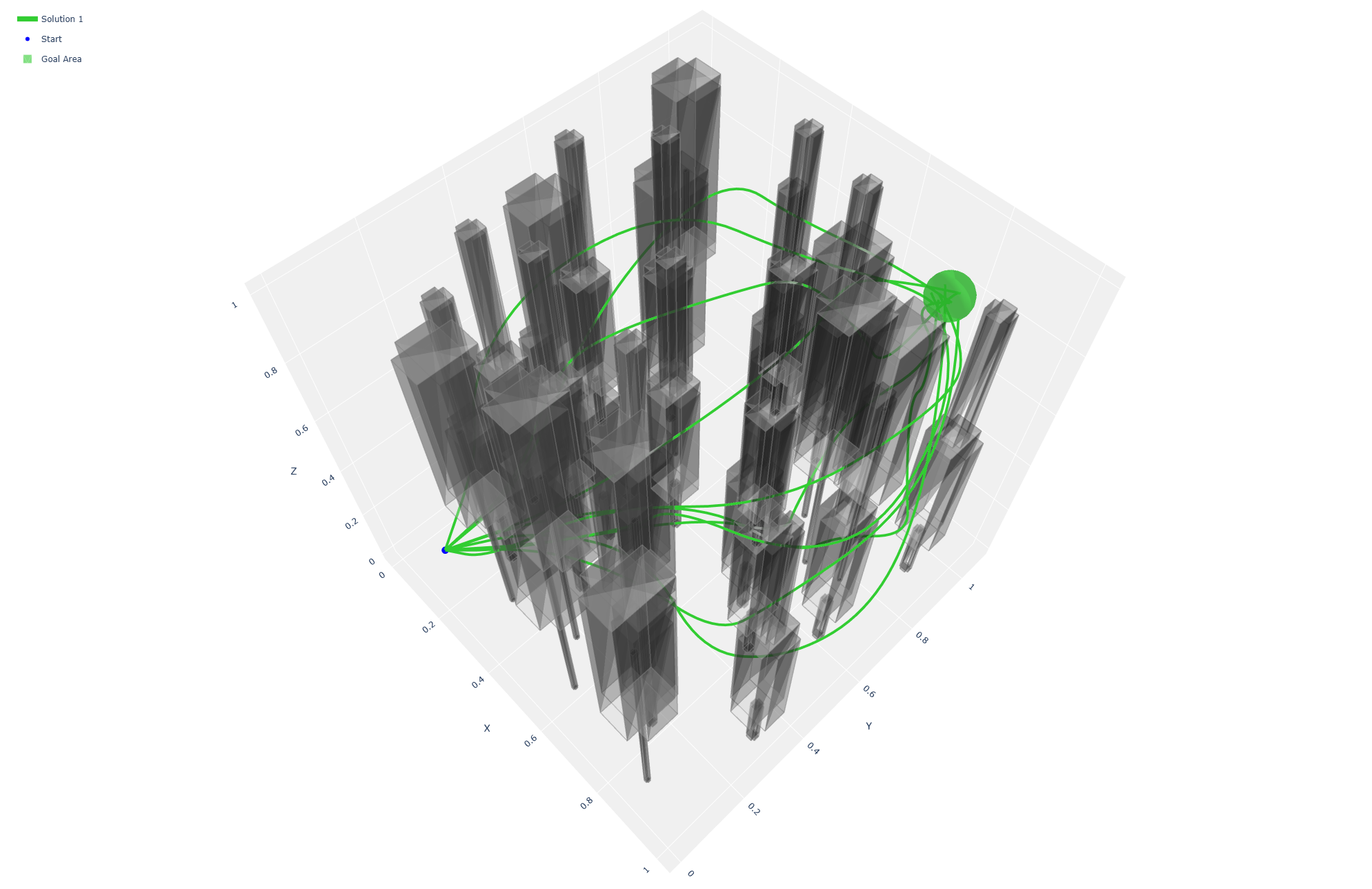}
         \caption{Initial Tree}
     \end{subfigure}
     \hfill
     \begin{subfigure}[b]{0.32\textwidth}
         \centering
         \adjincludegraphics[width=\textwidth, trim={0.1\width} 0 {0.1\width} 0, clip]{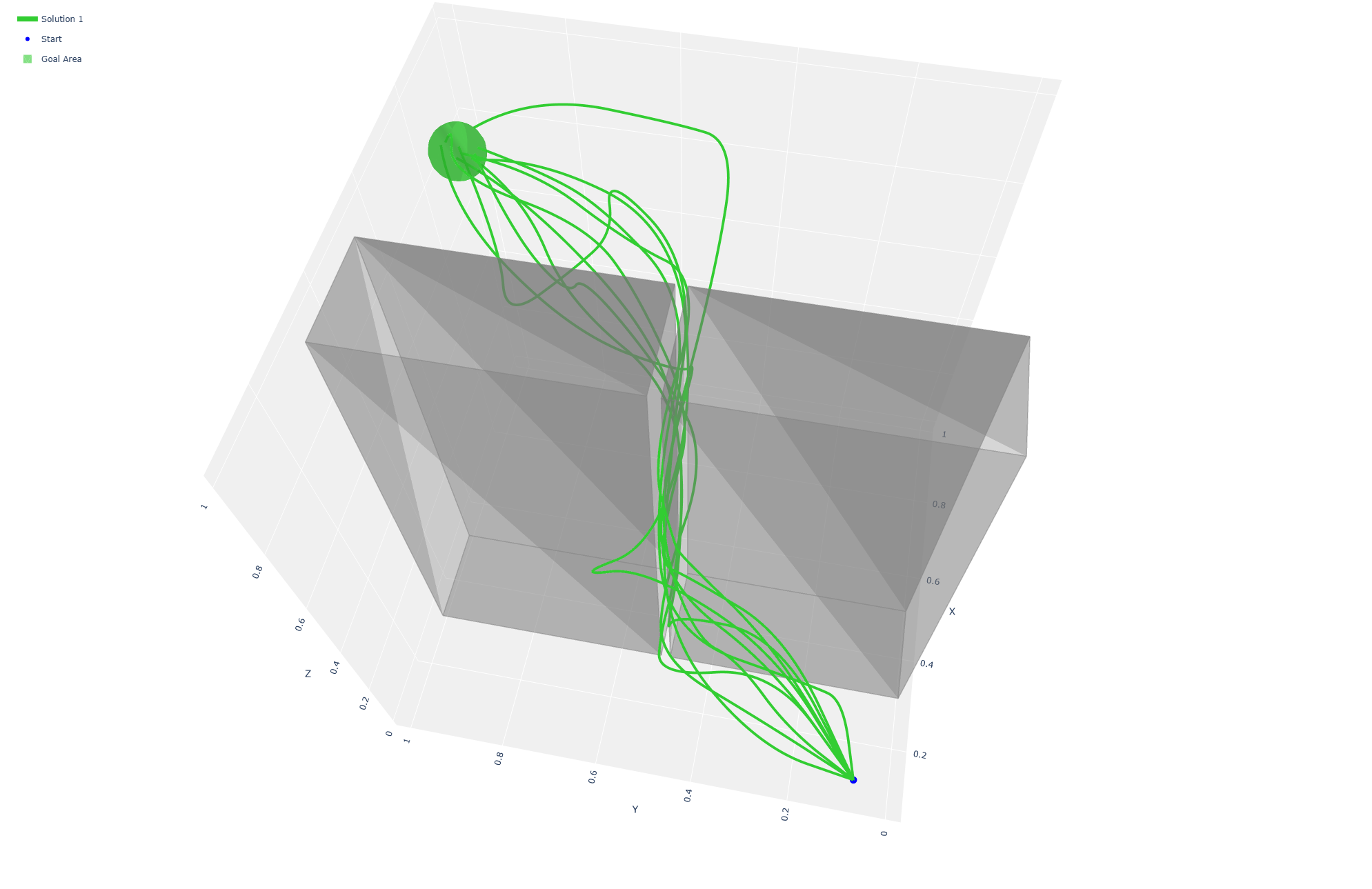}
         \caption{Initial Narrow}
     \end{subfigure}
     \hfill
     \begin{subfigure}[b]{0.32\textwidth}
         \centering
         \adjincludegraphics[width=\textwidth, trim={0.1\width} 0 {0.1\width} 0, clip]{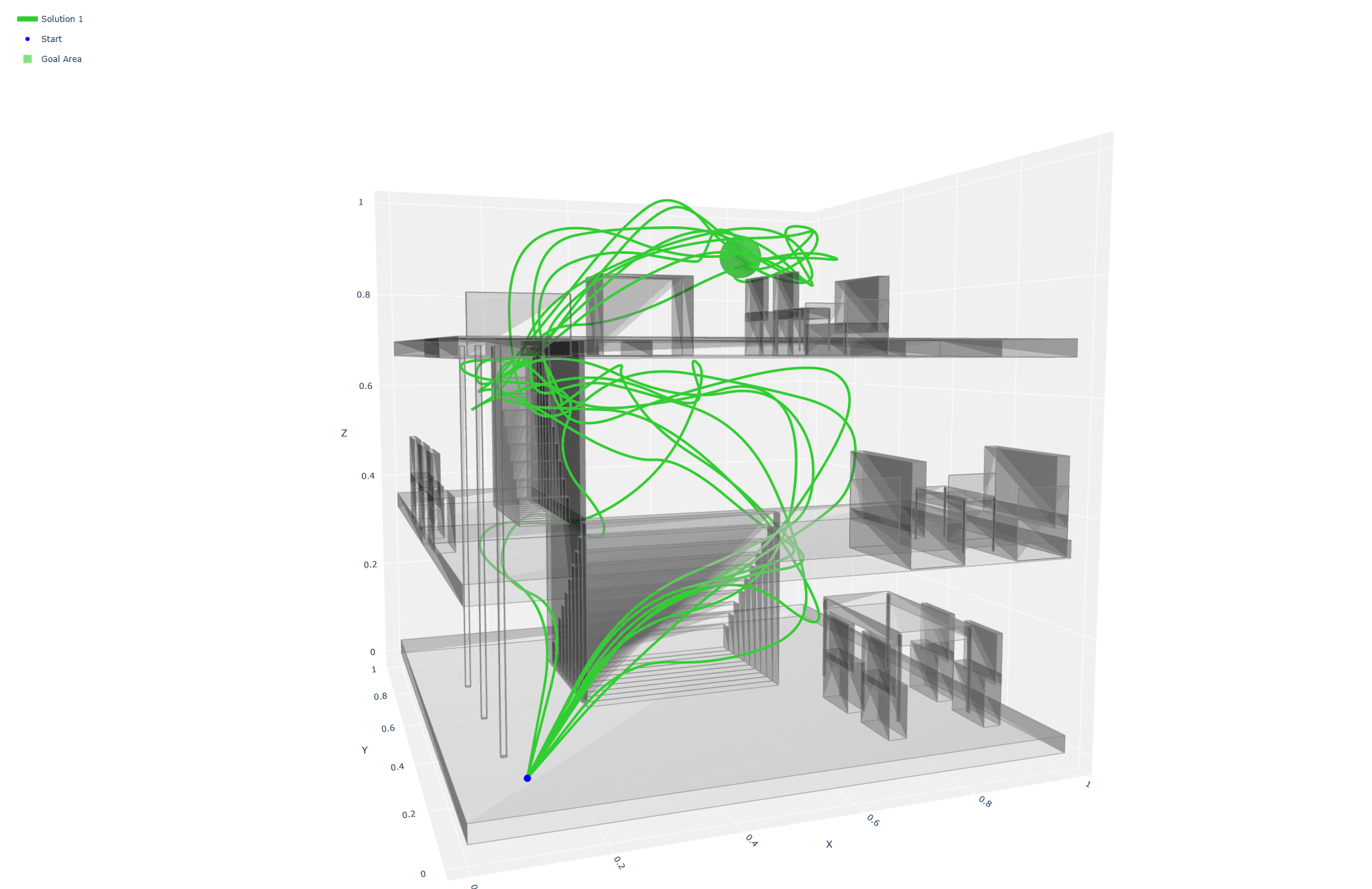}
         \caption{Initial House}
     \end{subfigure}

     \vspace{8pt} 

     \begin{subfigure}[b]{0.32\textwidth}
         \centering
         \adjincludegraphics[width=\textwidth, trim={0.1\width} 0 {0.1\width} 0, clip]{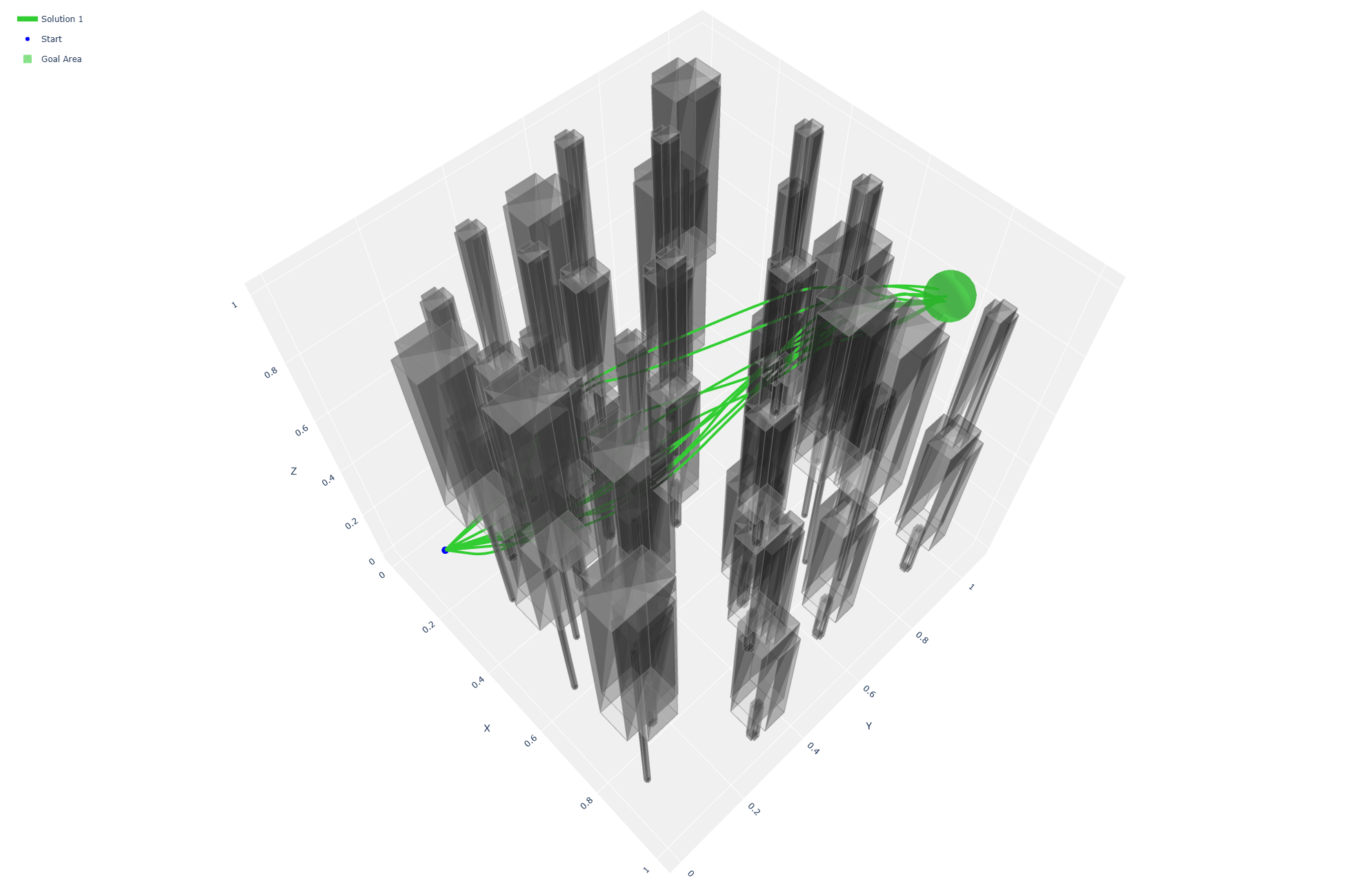}
         \caption{Final Tree}
     \end{subfigure}
     \hfill
     \begin{subfigure}[b]{0.32\textwidth}
         \centering
         \adjincludegraphics[width=\textwidth, trim={0.1\width} 0 {0.1\width} 0, clip]{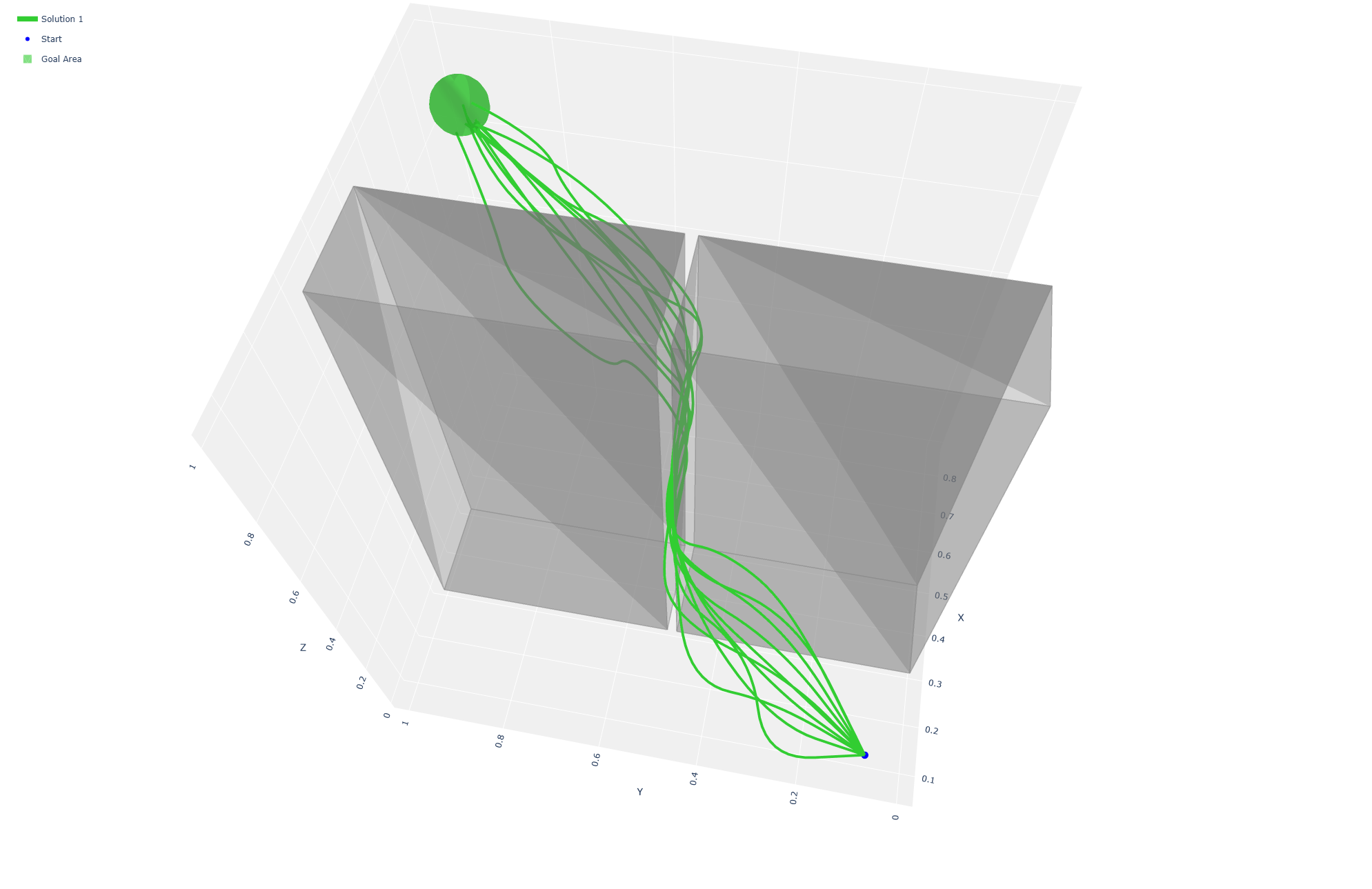}
         \caption{Final Narrow}
     \end{subfigure}
     \hfill
     \begin{subfigure}[b]{0.32\textwidth}
         \centering
         \adjincludegraphics[width=\textwidth, trim={0.1\width} 0 {0.1\width} 0, clip]{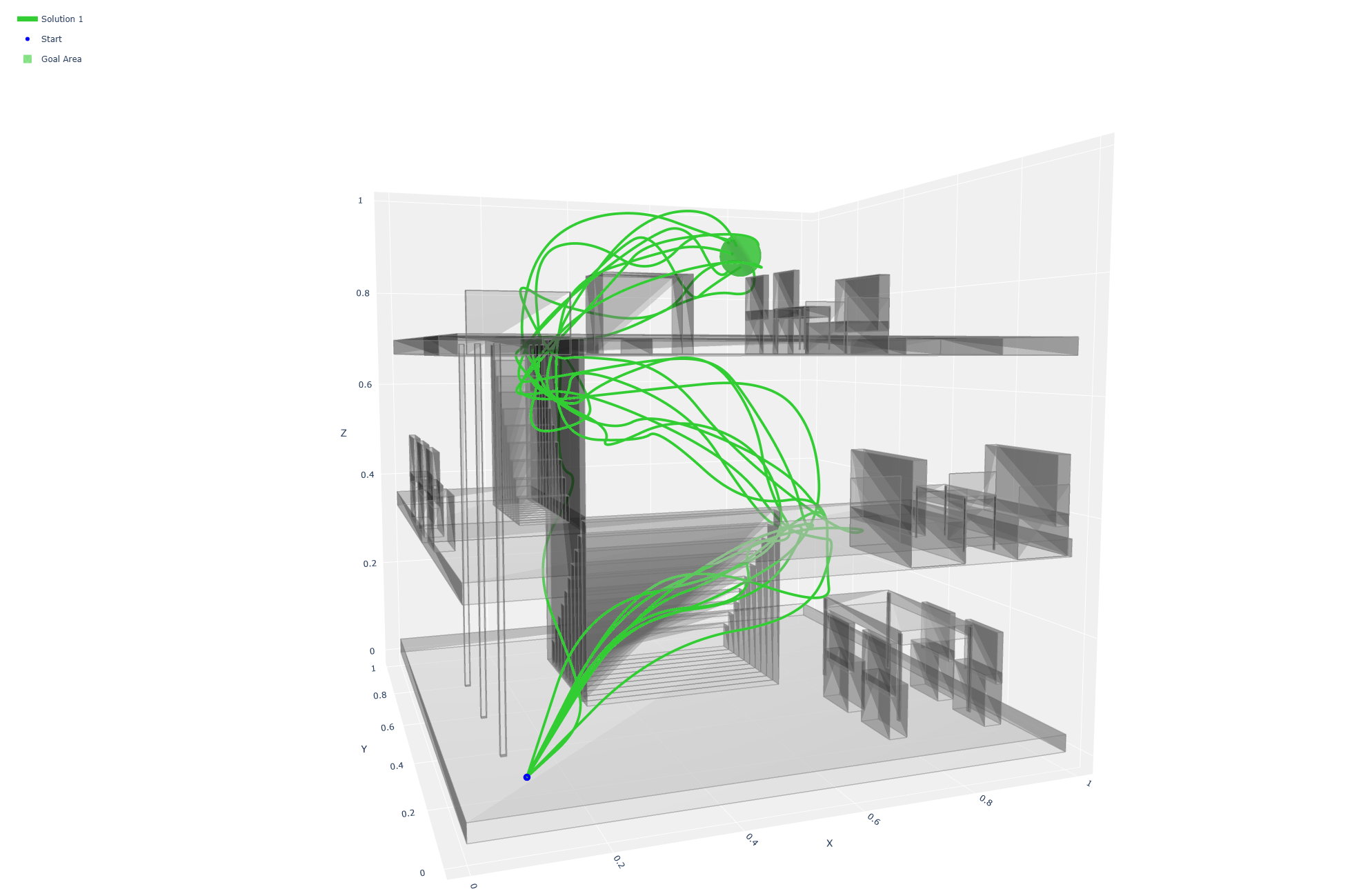}
         \caption{Final House}
     \end{subfigure}

     \caption{Depiction of 10 initial and final solution trajectories (green lines) and 3d obstacle environments for \methodname, run with the double integrator model. The start and goal regions are represented by the blue and green spheres. Cost criterion is set to distance traveled to better visualize convergence. Environments are taken from kino-PAX~\cite{kinopax}.}
     \label{fig:envs}
\end{figure*}

\subsection{Theoretical Analysis}

We establish that our approach is probabilistically complete and, through the AO-\textit{x} meta-algorithm, achieves asymptotic optimality under the condition that the cost function and robot dynamics are Lipschitz continuous.
We use a forward-propagation scheme to satisfy differential constraints. 
Our parallel kinodynamic RRT inherits probabilistic completeness from sequential kinodynamic RRT~\cite{pcrrt}. 
Our parallelization affects only the \emph{rate} of sampling by introducing a bounded delay on the discoverability of nodes, not the \emph{distribution} of samples.

Consider a batched planner with a branching factor $A=1$.
Unlike sequential kinodynamic RRT, where newly inserted nodes are immediately available for future nearest-neighbor queries, all nearest-neighbor queries within a batch are evaluated against a frozen snapshot of the tree. Consequently, the discovery of descendants of newly inserted nodes may be delayed by at most one batch. Probabilistic completeness of forward-propagation RRT relies on the existence of a finite sequence of successful extensions connecting the start state to the goal region, each occurring with non-zero probability~\cite{pcrrt}. Batching only requires successive extensions along such a sequence to occur in different batches, rather than different iterations, introducing a finite delay in node availability without changing the probability that each extension is eventually realized. Since the number of batches approaches infinity as the number of samples approaches infinity, this bounded delay does not affect asymptotic probabilistic completeness.

For a branching factor $A>1$, the planner performs $K=B/A$ nearest-neighbor queries and samples $A$ controls for each selected parent. Relative to a planner with batch size $K$ and a single control sample per parent, the branching planner evaluates the same set of parent nodes while augmenting each parent with an additional $A-1$ independently sampled controls. Thus, every candidate propagation considered by the $A=1$ planner is also considered by the branching planner, together with additional control samples. Since branching preserves all opportunities for successful expansion while introducing additional candidate propagations, it does not compromise probabilistic completeness.

Our use of a fixed-step propagator ($\Delta t$) for XLA-optimization purposes is consistent with the assumptions of completeness for discrete-time kinodynamic planners, provided $\Delta t$ is sufficiently small.

Since the inner planner is probabilistically complete and the cost
threshold $c_{\mathrm{best}}$ is monotonically decreasing across AO-x
iterations, the assumptions of AO-x~\cite{aox} remain
satisfied. Under Lipschitz continuous dynamics and cost functions, the
sequence of solutions produced by PAKR therefore converges almost surely
to the optimal cost $c^\ast$, establishing asymptotic optimality.

\section{Experiments}
\label{sec:exp}

We benchmark \methodname against Kino-PAX~\cite{kinopax}, another parallelized kinodynamic motion planner implemented in CUDA C, as well as the sequential CPU-based kinodynamic planners iDb-A*~\cite{idbastar}, and SST*~\cite{sst} from the Open Motion Planning Library (OMPL)~\cite{ompl}. We also show convergence results for \methodname, and explore the effects of batch size and branching factor on the performance of \methodname. Both \methodname and Kino-PAX are run on an NVIDIA GeForce RTX 4090 with 16,384 CUDA cores and 24 GB of VRAM, while iDb-A* and SST are executed on a 24-core AMD Ryzen Threadripper PRO 5965WX with a base clock speed of 3.8GHz and 32 GB of RAM. In all experiments, we report the median performance of each planner across 100 trials.

\subsection{Comparison With Kino-PAX}
\label{sec:kino-pax-benchmarks}
We evaluate the performance of \methodname and Kino-PAX on the same environments and dynamics in the original Kino-PAX paper:
\begin{enumerate}
    \item \textbf{Double Integrator (DI)} has a 6-dimensional state space $[x,y,z,\dot{x},\dot{y},\dot{z}]$ consisting of the 3D position and velocity, and a 3-dimensional control space $[\ddot{x},\ddot{y},\ddot{z}]$ consisting of the linear acceleration applied to each axis.
    \item \textbf{Dubins Airplane (DA)} has a 6-dimensional state space $[x,y,z,\phi,\theta,v]$ consisting of the 3D position, yaw, pitch, and forward speed, and a 3-dimensional control space $[\dot{\phi}, \dot{\theta}, \dot{v}]$ consisting of the yaw rate, pitch rate, and forward acceleration.
    \item \textbf{Quadcopter (QC)} has a 12-dimensional state space $[x,y,z,\phi,\theta,\psi,u,v,w,p,q,r]$ consisting of the 3D position, yaw, pitch, and roll angles, linear velocities in body frame, and angular velocities in body frame, and a 4-dimensional control space $[T, \tau_x, \tau_y, \tau_z]$ consisting of the thrust and torques in body frame.
\end{enumerate}
A depiction of the \textbf{Tree (A), Narrow (B), and House (C)} environments can be seen in \cref{fig:envs}. As Kino-PAX is not asymptotically optimal, we use a variant of \methodname without the AO-\textit{x} meta-algorithm, where a single parallel RRT iteration is run, for a fair comparison. Experimental results are shown in \cref{tab:results_comparison}. While there is an asymptotically near-optimal extension of Kino-PAX~\cite{perrault2026kino}, we do not evaluate against it due to the lack of an open-source implementation. In all environments and dynamics, \methodname consistently has a smaller tree size than Kino-PAX upon finding a solution, showing that it is more efficient at exploring the state space while achieving comparable runtime.

\begin{table}[b]
\centering
\caption{Performance comparison across environments (A, B, C) over 100 trials each. Success rates were 100\% for all trials. \methodname\ achieves comparable runtimes and consistently produces smaller tree sizes across all dynamics.}

\label{tab:results_comparison}
\footnotesize 
\addtolength{\tabcolsep}{-3.5pt} 
\begin{tabular}{@{}ll cc cc cc@{}}
\toprule
& & \multicolumn{2}{c}{\textbf{Env A}} & \multicolumn{2}{c}{\textbf{Env B}} & \multicolumn{2}{c}{\textbf{Env C}} \\
\cmidrule(lr){3-4} \cmidrule(lr){5-6} \cmidrule(lr){7-8}
\textbf{Dynamics} & \textbf{Planner} & \textbf{Time} & \textbf{Nodes} & \textbf{Time} & \textbf{Nodes} & \textbf{Time} & \textbf{Nodes} \\ 
\midrule

\multirow{2}{*}{\textbf{6D DI}} 
& KinoPAX & 5.0ms & 164k & 3.5ms & 144k & 6.3ms & 193k \\
& \methodname & \textbf{2.5ms} & \textbf{9k} & \textbf{1.8ms} & \textbf{21k} & \textbf{5.5ms} & \textbf{19k} \\ 
\midrule

\multirow{2}{*}{\textbf{6D DA}} 
& KinoPAX & \textbf{4.4ms} & 63k & \textbf{3.7ms} & 72k & \textbf{7.6ms} & 107k \\
& \methodname & 5.4ms & \textbf{19k} & 5.9ms & \textbf{32k} & 14.9ms & \textbf{57k} \\ 
\midrule

\multirow{2}{*}{\textbf{12D QC}} 
& KinoPAX & 17.8ms & 285k & 17.8ms & 285k & \textbf{24.6ms} & 333k \\
& \methodname & \textbf{14.9ms} & \textbf{71k} & \textbf{17.1ms} & \textbf{147k} & 51.4ms & \textbf{173k} \\ 
\bottomrule
\end{tabular}
\end{table}

\begin{table*}[t]
\centering
\caption{Convergence results for \methodname on Kino-PAX environments and models. The median times and costs of the initial and final solutions are recorded.}
\label{tab:convergence_stats}
\small
\setlength{\tabcolsep}{5.5pt} 
\begin{tabular}{@{}l cccc cccc cccc@{}}
\toprule
& \multicolumn{4}{c}{\textbf{Environment A}} & \multicolumn{4}{c}{\textbf{Environment B}} & \multicolumn{4}{c}{\textbf{Environment C}} \\ 
\cmidrule(lr){2-5} \cmidrule(lr){6-9} \cmidrule(lr){10-13}
\textbf{Dynamics} & Time$_1$ & Cost$_1$ & Time$_f$ & Cost$_f$ & Time$_1$ & Cost$_1$ & Time$_f$ & Cost$_f$ & Time$_1$ & Cost$_1$ & Time$_f$ & Cost$_f$ \\ \midrule

\textbf{6D DI} & 2.3ms & 1.76 & 28.0ms & 1.51 & 1.8ms & 2.03 & 20.0ms & 1.54 & 4.9ms & 3.11 & 45.0ms & 2.33 \\
\textbf{6D DA} & 4.8ms & 1.71 & 22.0ms & 1.56 & 5.9ms & 1.69 & 6.0ms  & 1.66 & 14.8ms & 3.42 & 32.0ms & 2.98 \\
\textbf{12D QC} & 13.1ms & 2.16 & 20.1ms & 1.98 & 17.1ms & 2.42 & 31.0ms & 2.11 & 55.6ms & 3.98 & 165.4ms & 3.62\\

\bottomrule
\end{tabular}
\end{table*}

\begin{figure}[htbp]
    \centering
    \begin{subfigure}{\linewidth}
        \centering
        \includegraphics[width=0.8\linewidth]{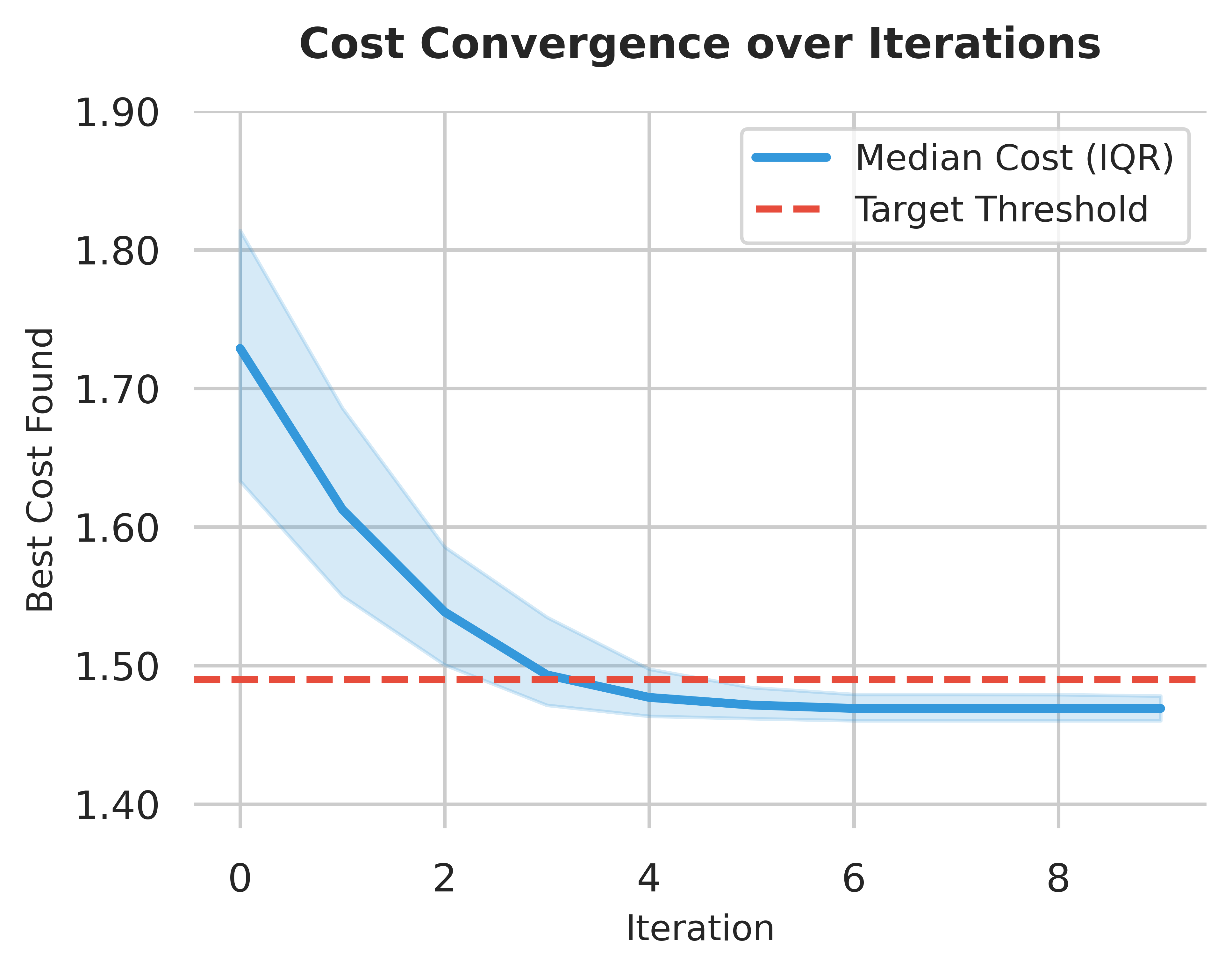}
    \end{subfigure}
    
    
    \begin{subfigure}{\linewidth}
        \centering
        \includegraphics[width=0.8\linewidth]{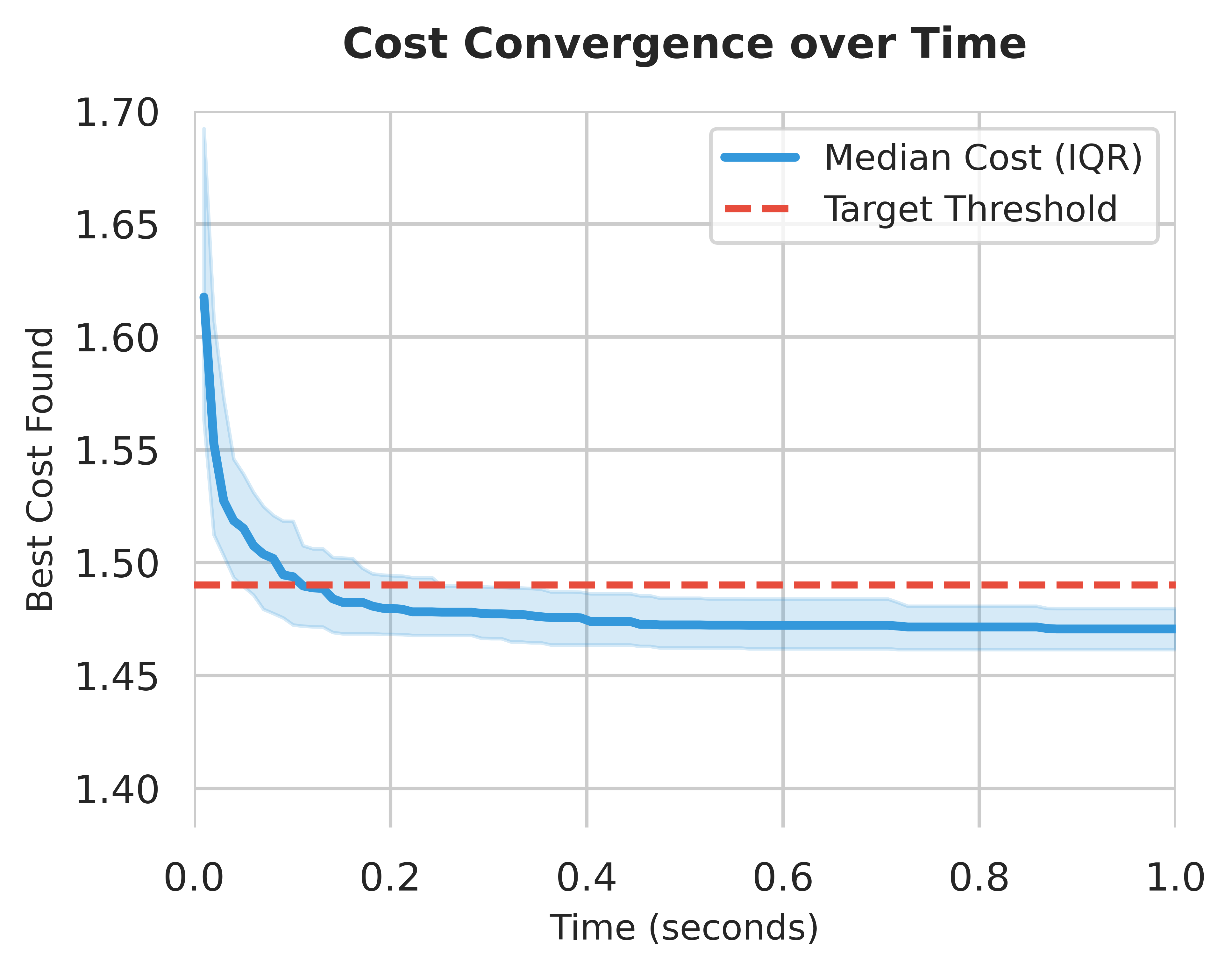}
    \end{subfigure}
    
    \caption{\methodname convergence analysis showing median cost and interquartile range (IQR) across 100 runs for the double integrator in environment A with a cost threshold of 1.49.}
    \label{fig:aorrt_analysis_combined}
\end{figure}

\subsection{AO Convergence}
We show the fast convergence of \methodname towards an optimal solution for the dynamics and environments described in \cref{sec:kino-pax-benchmarks}. A visualization of the initial and final trajectories can be found in \cref{fig:envs}. While it is standard for kinodynamic planners to find time-optimal trajectories using time duration as the cost criterion, we chose to measure distance traveled to make it easier to visualize the solution convergence. Numerical results are reported in \cref{tab:convergence_stats}, and \cref{fig:aorrt_analysis_combined} displays the convergence of the 6d double integrator in environment A relative to the number of iterations and time. We find that \methodname is able to significantly reduce the cost of solution trajectories in only tens of milliseconds.

\subsection{Comparison With iDb-A* and SST*}
\label{sec:dynobench-benchmarks}
\begin{figure} [b]
    \centering
    \includegraphics[width=0.8\linewidth]{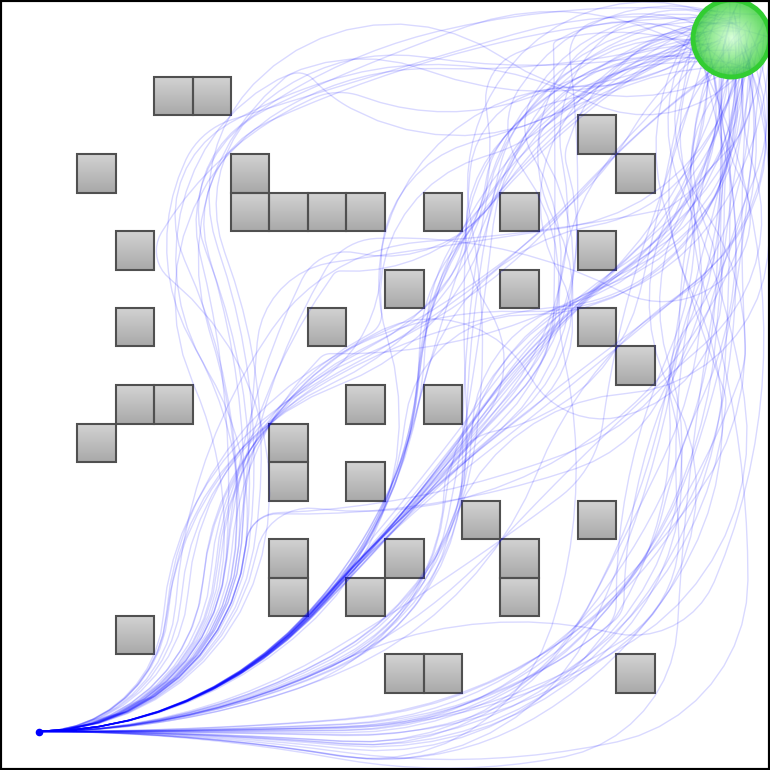}
    \caption{Illustration of 100 \methodname solution paths for Unicycle 1 using a randomly generated cluttered environment with 40 obstacles.}
    \label{fig:uni}
\end{figure}

To evaluate the quality of solutions produced, we compare the performance of \methodname with that of the asymptotically optimal iDb-A* and SST* planners on four dynamical systems from DynoBench, a benchmark for kinodynamic planning problems~\cite{idbastar}:
\begin{enumerate}
    \item \textbf{Unicycle 1} has a 3-dimensional state space $[x,y,\theta]$ consisting of the 2D position and heading, and a 2-dimensional control space $[v,\omega]$ consisting of the linear velocity in the direction of the heading and angular velocity.
    \item \textbf{Unicycle 2} has a 5-dimensional state space $[x,y,\theta,v,\omega]$ including the linear and angular velocities in the state, and a 2-dimensional control space $[a,\alpha]$ consisting of the linear and angular accelerations.
    \item \textbf{Acrobot} has a 4-dimensional state space $[\theta_1,\theta_2,\dot{\theta}_1,\dot{\theta}_2]$ consisting of the angles and angular velocities of the two links of double pendulum that is actuated at the second (elbow) joint, and has a 1-dimensional control space consisting of the torque applied at the second joint.
    \item \textbf{Quadcopter} has the same dynamics as the 12D Quadcopter described in \cref{sec:kino-pax-benchmarks}, and is referred to as ``Quadrotor v1" in DynoBench.
\end{enumerate}
For Unicycle 1 and Unicycle 2, the planners are evaluated on a custom problem with randomly distributed square obstacles, as illustrated in \cref{fig:uni}, while for Acrobot and Quadcopter the swing-up and window problems from DynoBench are used, respectively, for evaluation. More information about these systems and problems 
can be found in~\cite{idbastar}.

\begin{table}[b]
\centering
\caption{Benchmarks Against iDb-A* and SST*. Each planner was run on each problem for 100 trials, and the success rate, median initial solution time and cost, and final solution cost are shown.}
\label{tab:benchmark-dynoplan}
\small
\begin{tabular}{llrrrr}
\hline
Problem & Alg. & SR\% & $\text{Time}_1$ & $\text{Cost}_1$ & $\text{Cost}_f$ \\
\hline
\multirow{3}{*}{Unicycle 1}
 & iDb-A* & 100 & 254ms           & 5.80s           & 5.10s \\
 & SST*       & 100 & 600ms            & 8.34s          & \textbf{4.41s} \\
 & \methodname & 100 & \textbf{1.3ms}  & \textbf{4.62s}  & 4.46s \\
\hline
\multirow{3}{*}{Unicycle 2}
 & iDb-A* & 12  & 4637ms          & 8.70s           & 8.70s \\
 & SST*       & 100 & 398ms           & 14.15s          & 6.95s \\
 & \methodname & 100 & \textbf{1.4ms}  & \textbf{5.40s}  & \textbf{4.75s} \\
\hline
\multirow{3}{*}{Acrobot}
 & iDb-A* & 100 & 1450ms          & 5.24s           & 4.96s \\
 & SST*       & 74  & 3298ms          & 4.36s           & 3.39s \\
 & \methodname & 100 & \textbf{13ms}  & \textbf{3.87s}  & \textbf{3.22s} \\
\hline
\multirow{3}{*}{Quadcopter}
 & iDb-A* & 100 & 1387ms          & \textbf{2.56s}  & \textbf{2.21s} \\
 & SST*       & 41  & 85457ms         & 9.64s           & 9.42s \\
 & \methodname & 100 & \textbf{175ms}  & 7.54s           & 6.23s \\
\hline
\end{tabular}
\begin{tablenotes}
    \small
    \item Note: Unicycle 1 uses velocity control while Unicycle 2 uses force control.
\end{tablenotes}
\end{table}
Both iDb-A* and SST* are given a planning budget of 300 seconds to fairly evaluate their convergence towards the optimal solution. The hyperparameters of SST* (goal bias, selection radius, pruning radius) are also fine-tuned for each problem to achieve more competitive performance; additionally, for Acrobot and Quadcopter the goal region of SST* is expanded to increase likelihood of finding a solution, with an additional trajectory optimization step to find an exact solution. In line with DynoBench, the duration of the solution trajectory is used as the cost. Results are shown in \cref{tab:benchmark-dynoplan}. We find \methodname achieves orders of magnitude speed-up of initial solution time, as well as better initial solution cost in 3 out of 4 problems compared to iDb-A* and SST*, while also having better or competitive final solution cost. We believe that the particularly low success rate of iDb-A* on Unicycle 2 is due to its reliance on a trajectory optimization step to compute a feasible trajectory, which fails under the complex dynamics and cluttered environment.


\begin{table}[t]
\centering
\caption{Batch Size and Branching Factor ($A$) Sensitivity on 12D Quadcopter (Env A). Bolded values for global minimums for time, iteration count, and tree size.}
\label{tab:parameter_sweep}
\small 
\setlength{\tabcolsep}{3pt} 
\begin{tabular}{@{} l l cccc @{}}
\toprule
& & \multicolumn{4}{c}{\textbf{Branching Factor} ($A$)} \\ \cmidrule(l){3-6} 
\textbf{Batch} & \textbf{Metric} & \textbf{2} & \textbf{16} & \textbf{64} & \textbf{128} \\ \midrule

\multirow{4}{*}{\textbf{4k}} 
    & Success (\%) & 100.0 & 100.0 & 99.0 & 94.0 \\
    & Time (ms)    & 19.5  & 16.1  & 26.7 & 41.9 \\
    & Iters        & 22.0  & 36.0  & 69.0 & 104.0 \\
    & Nodes        & \textbf{37.6k} & 57.7k & 107.5k & 168.1k \\ \addlinespace

\multirow{4}{*}{\textbf{8k}} 
    & Success (\%) & 100.0 & 100.0 & 97.0 & 93.0 \\
    & Time (ms)    & 24.5  & \textbf{13.1}  & 19.0 & 26.7 \\
    & Iters        & 15.0  & 21.0  & 39.0 & 57.0  \\
    & Nodes        & 54.3k & 70.9k & 124.9k & 181.9k \\ \addlinespace

\multirow{4}{*}{\textbf{16k}} 
    & Success (\%) & 100.0 & 100.0 & 97.0 & 87.0 \\
    & Time (ms)    & 47.2  & 14.3  & 17.1 & 14.3 \\
    & Iters        & 12.0  & 14.0  & 25.0 & 25.0  \\
    & Nodes        & 89.9k & 103.8k & 162.7k & 163.3k \\ \addlinespace

\multirow{4}{*}{\textbf{32k}} 
    & Success (\%) & 100.0  & 100.0 & 99.0 & 86.0 \\
    & Time (ms)    & 115.3  & 26.9  & 20.1 & 19.9 \\
    & Iters        & \textbf{10.0}   & 12.0  & 17.0 & 20.0  \\
    & Nodes        & 154.6k & 175.5k & 232.3k & 277.0k \\ \bottomrule
\end{tabular}
\end{table}

\subsection{Effects of Batch Size and Branching Factor}
\label{sec:batch-branching}
We experiment with varying batch sizes and branching factors on the performance of \methodname with the 12D quadcopter in Environment A. Results are displayed in \cref{tab:parameter_sweep}. 

Due to the high dimensionality of the quadcopter system, a large branching factor is detrimental to exploration, causing success rates and runtime performance to worsen. However, a branching factor of 2 is tested to show its effect on reducing the nearest neighbor search overhead, especially for large batch sizes that approach the limits of GPU capacity. Both batch size and branching factor affect the exploration efficiency, which can be seen in the increasing tree size.

\subsection{Scalability to Heavy Simulation}
\label{sec:mjx-experiments}

\begin{figure} [t]
    \centering
    \includegraphics[width=0.8\linewidth]{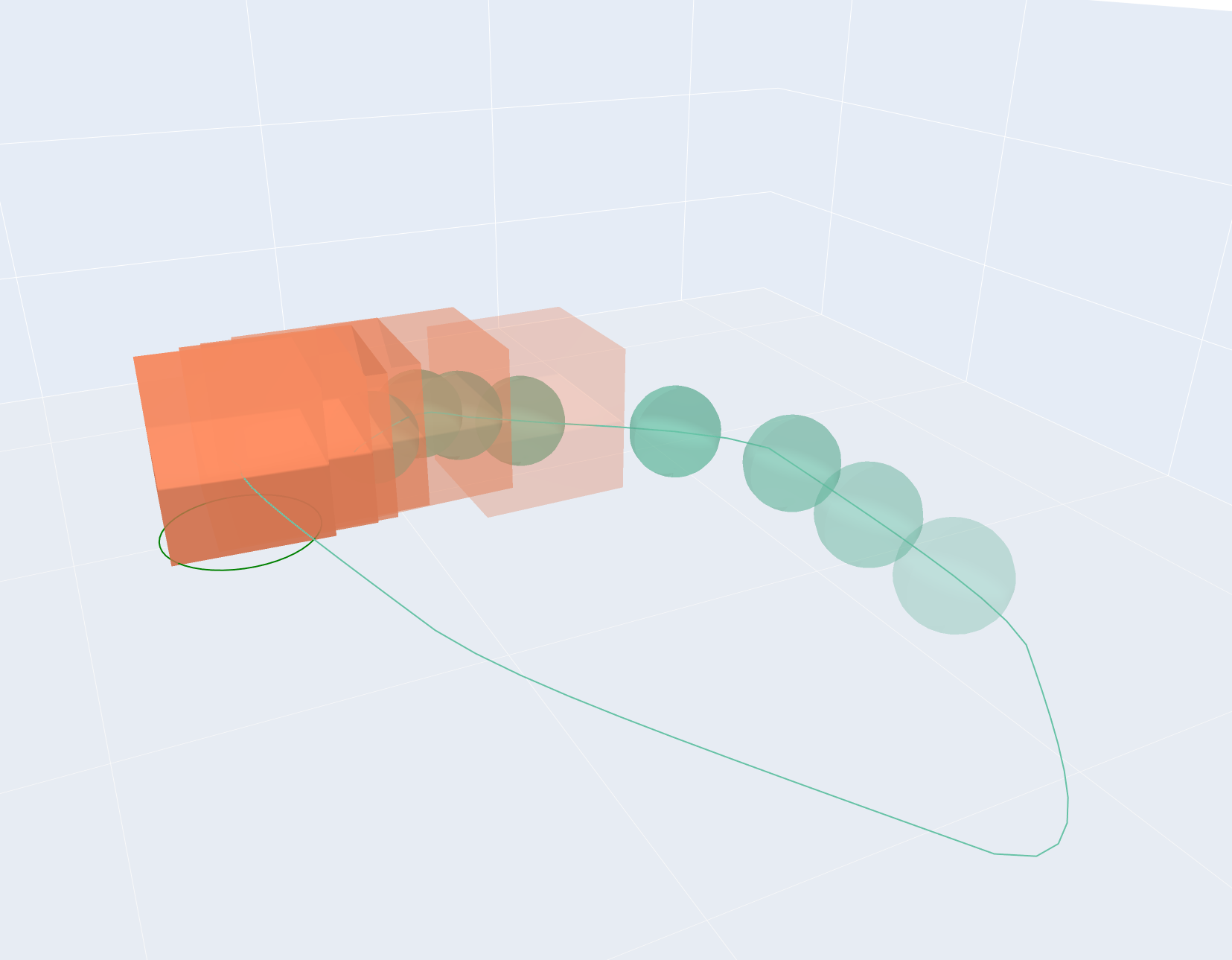}
    \caption{Solution trajectory for the 10-DoF block push problem. Earlier frames are shown with a lower opacity.}
    \label{fig:ee}
\end{figure}

\begin{figure} [t]
    \centering
    \includegraphics[width=0.8\linewidth]{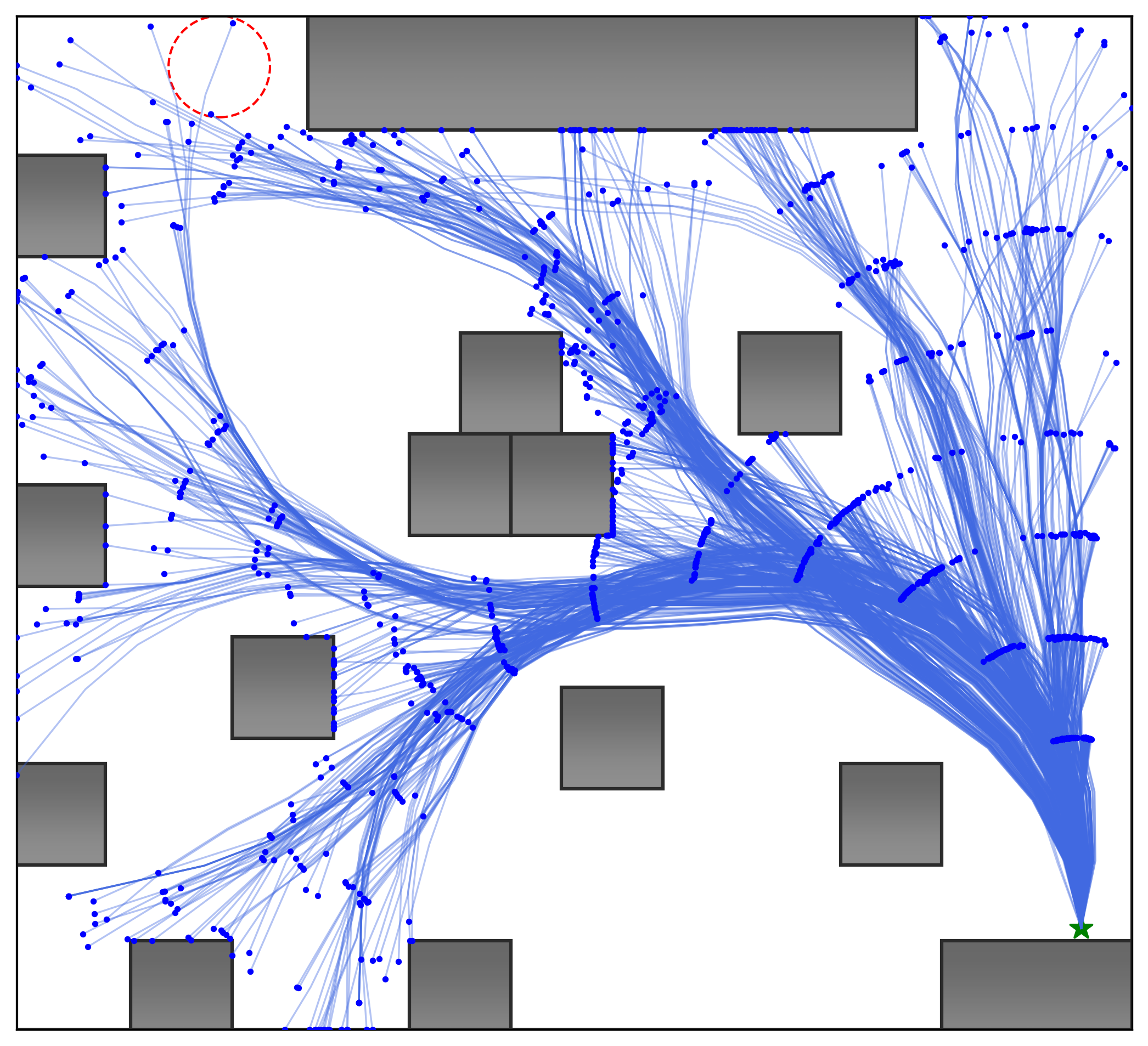}
    \caption{Illustration of \methodname's search tree for the soft robot~\cite{vine}. 
    A solution is found in a median time of 360.25ms, compared to the original paper's result on the order of seconds.}
    \label{fig:vine}
\end{figure}

To show \methodname's ability to scale to complex physical simulation, we demonstrate our approach on Cartpole and Block Push problems using Mujoco XLA~\cite{mujoco}, as well as a soft growing vine robot planning problem using the soft robot simulator from Gao et al.~\cite{vine}:

\begin{enumerate}
    \item \textbf{Cartpole} has a 4-dimensional state space $[x, \theta,v,\omega]$ and a 1-dimensional control space $[a]$ for the cart's linear acceleration.
    \item \textbf{Block Push} is defined by a 10-dimensional state space $\mathbf{s} \in \mathbb{R}^{10}$, which can be decomposed into the configuration vector $\mathbf{q}$ and the velocity vector $\dot{\mathbf{q}}$:$$\mathbf{s} = \begin{bmatrix} \mathbf{q} \\ \dot{\mathbf{q}} \end{bmatrix}, \quad \text{where} \quad 
    \mathbf{q} = \begin{bmatrix} x_{ee} & y_{ee} & x_{b} & y_{b} & \theta_{b} \end{bmatrix}^\top$$
    We simplify the standard robot manipulator problem by reducing the state space, only accounting for the end effector position and velocity instead of the $q$ and $\dot{q}$ of the entire arm. We also restrict the end effector and block positions to the 2d plane. The controls $[\ddot{x}_{ee}, \ddot{y}_{ee}]$ are forces acting upon the end effector.
    \item \textbf{Vine}~\cite{vine} decomposes a soft growing vine robot into a multi-body link, calculating the joint angles and the length of the distal tip as the vine grows and actuation is applied to the body. While it keeps track of all joints, the planner only considers the tip position $[x, y, \theta]$ for its distance calculations. Its action space $[p, l_0]$ are two pneumatic actuator inputs that determine the bending moment at each joint.
\end{enumerate}

 \cref{fig:ee} shows a trajectory produced by \methodname for the block push problem, while \cref{fig:vine} displays the resultant tree for the vine planner. The median times to find a solution for Cartpole, Block Push, and Vine were 22.82, 283.65, and 360.25 ms, respectively. This increase in runtime can be attributed to heavier computation during propagation when using a physics simulator instead of analytical functions.

\section{Conclusion and Future Work}
\label{sec:disc}

We present \methodname, a high-performance kinodynamic motion planner that balances rapid execution with high solution quality. By parallelizing the AO-\textit{x} meta-algorithm with a fast internal RRT-based approach, we transform fast, sub-optimal sampling into an asymptotically optimal process through iterative, cost-bounded replanning. Our implementation uses JAX to fuse the entire planning loop into a single GPU kernel, eliminating CPU-GPU overhead and making high-speed kinodynamic planning accessible through standard Python tooling.

In summary, our contributions include: a massively parallel sampling-based motion planner utilizing XLA compilation for efficient GPU execution, an adaptation of the AO-\textit{x} algorithm that achieves asymptotic optimality in a parallelized setting, demonstrated scalability across complex environments, including high-dimensional MuJoCo-XLA domains, and experimental validation showing that \methodname achieves competitive or superior performance compared to state-of-the-art planners, frequently finding solutions in milliseconds.

Future work includes extending this planner to more difficult, higher-dimensional problems such as non-prehensile manipulation with a 7-DoF arm, integrating learned action proposals rather than random sampling, and evaluating against Kino-PAX$^+$~\cite{perrault2026kino} when available.


\bibliographystyle{IEEEtran}
\bibliography{references}

\end{document}

%% file: preamble.tex
\usepackage[utf8]{inputenc}

\usepackage{comment}
\usepackage{xspace}
\usepackage{graphicx}
\usepackage{overpic}
\usepackage{svg}
\usepackage{xcolor}
\usepackage{bm}
\usepackage[hang,flushmargin]{footmisc}
\usepackage{adjustbox}

\usepackage{tabularx}
\usepackage{multirow}
\usepackage{booktabs}
\usepackage{colortbl}
\usepackage{float}
\usepackage{placeins}
\usepackage{xspace}

\usepackage{array}
\newcolumntype{P}[1]{>{\centering\arraybackslash}p{#1}}  

\usepackage{etoolbox}
\makeatletter
\patchcmd{\@makecaption}
{\scshape}
{}
{}
{}
\newcommand{\onetagright}{\tagsleft@false}
\makeatother

\usepackage{amsmath}
\usepackage{amssymb}
\usepackage{amsthm}
\usepackage{siunitx}

\newtheoremstyle{main}
{1em}                                                
{1em}                                                
{\itshape}                                        
{0pt}                                                
{\scshape}                                           
{\\*}                                                
{2pt}                                                
{\thmname{#1}\thmnumber{ #2}: \thmnote{\itshape #3}} 

\usepackage{environ}

\usepackage[linesnumbered,ruled,noend]{algorithm2e}
\newcommand{\removelatexerror}{\let\@latex@error\@gobble}

\usepackage[
	activate   = {true},
	protrusion = false,
	expansion  = true,
	kerning    = true,
	spacing    = true,
	tracking   = false,
	auto       = true,
	selected   = true,
	factor     = 1000,
	stretch    = 10,
	shrink     = 10,
]{microtype}

\definecolor{purduegold}{HTML}{C28E0E} 

\makeatletter
\let\NAT@parse\undefined
\makeatother
\usepackage[pdfa,colorlinks,bookmarksopen,bookmarksnumbered,allcolors=purduegold]{hyperref}
\usepackage[english]{babel}

\makeatletter
\newcommand\footnoteref[1]{\protected@xdef\@thefnmark{\ref{#1}}\@footnotemark}
\makeatother

\usepackage[nameinlink,capitalise]{cleveref}
\crefname{line}{line}{lines}
\crefname{figure}{Fig.}{Figs.}
\Crefname{figure}{Fig.}{Figs.}
\crefname{equation}{Eq.}{Eqs.}
\Crefname{equation}{Eq.}{Eqs.}
\crefname{section}{Sec.}{Secs.}
\Crefname{section}{Sec.}{Secs.}
\crefname{definition}{Def.}{Defs.}
\Crefname{definition}{Def.}{Defs.}
\crefname{algorithm}{Alg.}{Algs.}
\Crefname{algorithm}{Alg.}{Algs.}
\crefname{assumption}{Asm.}{Asms.}
\Crefname{assumption}{Asm.}{Asms.}
\crefname{subassumption}{Asm.}{Asms.}
\Crefname{subassumption}{Asm.}{Asms.}
\Crefname{problem}{Problem}{Problems}
\crefname{problem}{Problem}{Problems}

\usepackage{pbalance}
\usepackage{flushend}

\let\labelindent\relax
\usepackage[inline]{enumitem}
\usepackage{mathtools}
\usepackage[normalem]{ulem}

\graphicspath{ {./images/} }

\newcommand{\methodname}{PAKR\xspace}

\usepackage{subcaption}
\usepackage{threeparttable}